%% file: main.tex
\documentclass[twoside]{article}

\usepackage[accepted]{styles/aistats2022}

\usepackage{styles/user_commands}
\usepackage{styles/user_packages}

\begin{document}
\runningtitle{CATVI for Hierarchical Bayesian Nonparametric Models}

\twocolumn[

\aistatstitle{CATVI: Conditional and Adaptively Truncated Variational Inference for Hierarchical Bayesian Nonparametric Models}

\aistatsauthor{ Yirui Liu \And Xinghao Qiao \And  Jessica Lam }

\aistatsaddress{ London School of Economics \And  London School of Economics \And JP Morgan Chase \& Co. } ]

\begin{abstract}
Current variational inference methods for hierarchical Bayesian nonparametric models can neither characterize the correlation structure among latent variables due to the mean-field setting, nor infer the true posterior dimension because of the universal truncation. To overcome these limitations, we propose the conditional and adaptively truncated variational inference method (CATVI) by maximizing the nonparametric evidence lower bound and integrating Monte Carlo into the variational inference framework. CATVI enjoys several advantages over traditional methods, including a smaller divergence between variational and true posteriors, reduced risk of underfitting or overfitting, and improved prediction accuracy. Empirical studies on three large datasets reveal that CATVI applied in Bayesian nonparametric topic models substantially outperforms competing models, providing lower perplexity and clearer topic-words clustering. 
\end{abstract}

\section{INTRODUCTION}
\label{sec:intro}
Hierarchical Bayesian nonparametric (HBNP) models are widely used in bioinfomatics, language processing, computer vision and network analysis \citep{sudderth2009, caron2017, williamson2016, yurochkin_bayesian_2019}. A major benefit of HBNP models is their ability to relax the fixed dimension assumption in parametric models. For example, in natural language processing, hierarchical Dirichlet process (HDP) model \citep{teh2006} replaces the finite-dimensional Dirichlet distribution in latent Dirichlet allocation (LDA) with a countable-dimensional Dirichlet process (DP). This is done by regarding the number of topics as a random variable that can be inferred from the data, rather than as a parametric value \citep{blei2003}.

However, it is much harder to implement HBNP models than their parametric counterparts. In particular, due to a HBNP model's infinite-dimensional nature, a finite-dimensional truncation is needed to approximate the posterior. Yet, the selection of the optimal truncation level poses several challenges. On one hand, the traditional Markov chain Monte Carlo (MCMC) methods \cite[]{teh2006} can produce an adaptive selection of the truncated dimension, but they are not computationally scalable especially for big data. 
On the other hand, standard variational inference methods \cite[]{teh2008, wang2011, hoffman2013, roychowdhury2015, xu_variational_2019} can accelerate the computation, but they suffer from a universal selection of the truncation level by truncating the dimension of all latent variables to a prespecified value. Using a prespecified value is problematic, because a subjective selection of the fixed truncation level can make the model prone to overfitting or underfitting, leading to low predictive accuracy. These existing challenges in universal truncation contradict the motivation and advantages of using HBNP models.

In this paper, we propose a general framework, called conditional and adaptively truncated variational inference (CATVI), to infer HBNP models in the following steps.
First, we convert the inference problem to an optimization task of maximizing our proposed nonparametric evidence lower bound based on finite partitions.
Second, we introduce a conditional setting when factorizing variational distributions by conditioning variables in the middle layers on two adjacent layers. Third, to handle big data, we develop a stochastic variational inference framework \cite[]{blei2017} under our conditional setting. Finally, we obtain empirical distributions from Monte Carlo sampling of local latent variables, which are then used to update the variational parameters for the global latent variables. This enables us to truncate the dimension of the latent variational distributions to that of the empirical distribution.

Our proposed method benefits from both the inferential accuracy of Monte Carlo sampling and the computational efficiency of variational inference. 
First, our method rebuilds the correlation structure and hence attains a smaller Kullback--Leibler (KL) divergence between the variational distribution and the true posterior.
Such procedure removes the unrealistic mean-field assumption, and searches for an optimal variational distribution over a wider family. Second, it adjusts the dimension of variation distributions, which converges to a stable level balancing the goodness-of-fit and model complexity. With these advantages, CATVI provides an adaptive selection of the truncated dimension, reducing the risk of overfitting or underfitting, while also enabling more accurate predictions without sacrificing the computational efficiency. Specific to the inference for the HDP model, CATVI enjoys several advantages over existing methods \citep{teh2006, hoffman2013, wang2012, bryant2012}, see Section~\ref{sec:comparision} for our detailed discussion. 

\section{BACKGROUND: HBNP MODELS}
\label{sec:model}
\begin{figure}
    \centering
    \includestandalone[width=0.8\linewidth]{graphic_model}
    \caption{The HBNP models. The blue and red boxes correspond to $J$ and $N_j$ replicates, respectively. }
    \label{fig:pic1}
    \vskip -0.1in
\end{figure}
As a subclass of Bayesian nonparametric models, HBNP models extend the simplicity of using random measures (see Appendix~\ref{app:CRM}) as priors to the following hierarchical structure, 
\begin{equation}
\label{equ:model}
\begin{gathered}
    G_0 | H \sim P(H),  ~~\beta | \lambda \sim p(\beta | \lambda), ~~
    G_j | G_0 \sim R(G_0),  
    \\
    ~~
    z_{ji} | G_j \sim G_j, ~~ x_{ji} |  z_{ji} \sim f(x_{ji} | \beta,z_{ji}),
\end{gathered}
\end{equation}
for $j =  1,\dots, J, i =  1,\dots, N_j,$ as illustrated in Figure~\ref{fig:pic1}. In the top layer,  $G_1, \dots, G_J$ are generated from a random measure $R$ with common base measure $G_0$, while in the bottom layer, $G_0$ itself is a realization of random measure $P$ with base measure $H$. To ensure exchangeability, $G_1, \dots, G_J$ are assumed to be identical and independent given $G_0$. Each local latent variable $z_{ji}$ is sampled from $G_j$ independently. Finally, the global parameter $\beta$ is assigned a prior $p(\beta | \lambda)$, and the observation $x_{ji}$ is generated from a likelihood function $f,$ parameterized by both global latent variable $\beta$ and local latent variables $z_{ji}$.

In topic modelling, the HDP model \citep{teh2006} uses a DP for both $P$ and $R$ in (\ref{equ:model}) as,
\begin{equation}
\begin{gathered}
\label{equ:teh-hdp}
    G_0 | H  \sim \DirPro (\alpha H), ~~
    G_j | G_0 \sim \DirPro (\gamma G_0),
\end{gathered}
\end{equation}
where $\alpha, \gamma$ are concentration parameters, and $H, G_0$ are normalized based measures (see Appendix~\ref{app:CRM}).
Suppose a corpus has $J$ documents, each document $j$ has $N_j$ words, and each word is chosen from a vocabulary with $W$ terms. Specifically, $G_0= \sum_{k=1}^{\infty} G_{0k}\delta_{\phi_k}$ is generated from the distribution $\DirPro(\alpha H)$, and 
for each document $j$, a topic proportion, defined as $G_j = \sum_{k=1}^{\infty} G_{jk}\delta_{\phi_k}$, is independently sampled from the distribution $\DirPro(\gamma G_0)$. For each topic $k,$ the distribution of words over vocabulary is sampled from a $W$-dimensional Dirichlet distribution parameterized by $\eta$, $\beta_k =(\beta_{k,1},\cdots, \beta_{k,W})^{\T}\sim  \text{Dir}(\eta)$. For each word $i$ in document $j$, a topic assignment $z_{ji}= \phi_k$ is chosen from $z_{ji} \sim \text{Multinomial}(G_j)$, where $\phi_k$ represents topic $k$. Finally, the observation $x_{ji}$ is generated from the assigned topic and the corresponding within-topic word distribution, $x_{ji} | \{z_{ji}=\phi_k\} \sim \text{Multinomial}(\beta_k)$.

The necessity to let $G_0$ be atomic can be shown in the HDP model. If $G_1,\dots, G_J$ are sampled from a Dirichlet process with a diffuse base measure instead of an atomic $G_0$, $G_1,\dots, G_J$ will not share any support almost surely, and thus none of the topics being shared across the documents. However, for a general HBNP model, as long as $G_0$ is atomic, it is not necessary to restrict the prior for $G_0$ to be a Dirichlet process or a probability random measure. For example, the $\Gamma$DP model, which has the following structure,
\begin{equation}
\begin{gathered}
\label{equ:gamma-dp-framework}
G_0 | H  \sim \Gamma \text{P} (\alpha H),  ~~
G_j | G_0 \sim \DirPro (G_0),  
\end{gathered}
\end{equation}
allows for more flexibility by removing the constraint on the concentration parameter in the top layer. Other choices of the prior for $G_0$ include beta process, stable process and inverse Gaussian process \citep{ghosal2017}. 

To infer the HBNP models, we set up the theoretical foundations for nonparametric KL divergence and evidence lower bound, and then propose a novel methodology in the following Sections~\ref{sec:cvi} and \ref{sec:method}.

\section{NONPARAMETRIC EVIDENCE LOWER BOUND}
\label{sec:cvi}
\subsection{KL Divergence between Random Measures}
The object of variational inference is to minimize the KL divergence between the variational distribution and the true posterior. For two infinite-dimensional random measures, their KL divergence is well defined even though an infinite-dimensional density function does not exist in a conventional sense.
Given two random measures $P$ and $Q$ from $(\Theta, \cM)$ into $(\Omega, \cF)$, 
the Radon--Nikodym derivative $dQ/dP$ exits if $Q$ is absolutely continuous with respect to $P$. Their KL divergence is defined as
$$
\label{equ:K-L_Process}
\KL(Q \parallel P) =  \int_{\Theta}    \log (dQ/dP) dQ,
$$
which is intractable due to the infinite-dimensional integral \citep{matthews2016}. We have developed a new approach to calculate it using the limit superior of the divergence between corresponding finite-dimensional induced measures, that is,
\begin{equation}
\label{equ:5}
\KL(Q \parallel P)
= \limsup_{\vO } \KL (q^{\vO } \parallel p^{\vO}),
\end{equation}
where $p^{\vO}$ and $q^{\vO}$ are respectively induced measures from $P$ and $Q$ on a finite partition $\varOmega=(A_1, \dots, A_n),$ such that $p^{\vO}(A_i) = P(A_i)$ and $q^{\vO}(A_i) = Q(A_i)$ for each $A_i \in \varOmega$. With an induced random variable $Z^\vO \colon \Theta \to \R^n$, we can also denote the induced measures by $p(Z^\vO)$ and $q(Z^\vO)$.
The result in (\ref{equ:5}) is justified in Appendix~\ref{app:induced}. 
We use the following two examples to illustrate (\ref{equ:5}).

\paragraph{Example 1}
For Poisson processes $P=\text{PP}(\Lambda + b \delta_\phi)$ and $Q=\text{PP}(\Lambda + a \delta_\phi)$, where $\Lambda$ is the intensity function defined on $\Omega$, $a, b \in \R^+$,  and $\delta_{\phi}$ is a Dirac function at point $\phi \in \Omega$. Under partition $\varOmega = (\phi, \Omega/\phi),$ the limit superior in (\ref{equ:5}) is achieved, that is,
\begin{equation*}
    \KL (Q \parallel P) =
    \KL \big(\text{Pois}(a) \parallel \text{Pois}(b)\big),
\end{equation*}
where $\text{Pois}(a)$ is the Poisson distribution with intensity $a$. 

\paragraph{Example 2} 
For Dirichlet processes $P=\DirPro(\alpha H+ \sum_{i=1}^{n} b_i \delta_{\phi_i})$ and $Q=\DirPro(\alpha H +\sum_{i=1}^{n} a_i \delta_{\phi_i})$, where $H$ is the base measure, $\alpha$ is the concentration parameter, $a_i, b_i \in \R^+$ and $\phi_i \in \Omega $ for $i=1, \dots n.$ Similarly, under the partition $\varOmega = \big(\phi_1, \dots, \phi_n,  \Omega/\{\phi_i\}_{i=1}^n\big),$ 
\begin{equation*}
\begin{gathered}
    \KL (Q \parallel P) = \\
    \KL \big(\text{Dir}(\alpha, a_1, \dots, a_n) \parallel \text{Dir}(\alpha, b_1,\dots, b_n)\big),
\end{gathered}
\end{equation*}
where $\text{Dir}(\alpha, a_1, \dots, a_n)$ is the Dirichlet distribution with parameters $\alpha, a_1, \dots, a_n$.

With the KL divergence between random measures represented under a finite partition, we can then define the nonparametric counterpart of evidence lower bound below.

\subsection{Nonparametric Evidence Lower Bound}
The parametric variational inference algorithm uses a finite-dimensional variational distribution to approximate the posterior by maximizing the evidence lower bound \citep{blei2017}. In contrast, HBNP models uses a random measure for the variational distribution, due to the infinite dimensionality of latent variables.
We propose a general inference framework for HBNP models by maximizing the nonparametric evidence lower bound (NPELBO), defined as the limit inferior of the parametric evidence lower bound under a finite partition, $\liminf_{\vO} (\text{ELBO}^\vO),$ that is
\begin{equation} 
\label{eq:ELBO}
\begin{gathered}
\liminf_{\vO } \big\{ \E_{q(Z^\vO)} \log p(X, Z^\vO) - 
\E_{q(Z^\vO) }\log q(Z^\vO) \big\},
\end{gathered}
\end{equation}
where $p(X, Z^\vO)$ and $q(Z^\vO)$ correspond to the induced measures from the joint distribution and the variational distribution on $\varOmega$, and where $Z$ and $X$ are the observations and latent variables, respectively.
Moreover, given the KL divergence between random measures in (\ref{equ:5}), in Appendix~\ref{app:induced_EBLo} we show that
\begin{equation}
\label{equ:7}
\KL \big(Q (Z) \parallel P(Z | X)\big) 
+ \text{NPELBO} = \log p(X).
\end{equation}
This demonstrates the equivalence between maximizing the NPELBO in (\ref{eq:ELBO}) and minimizing the KL divergence between the variational distribution $Q(Z)$ and the true posterior $P(Z | X).$ 
The task of maximizing the NPELBO is general and can be applied broadly within Bayesian nonparametrics. To simplify notation, we will use $p(\cdot)$ and $q(\cdot)$ to denote the true and variational distributions, respectively, where the context is clear. To infer HBNP models, we aim to maximize the defined NPELBO, while truncating the dimension of variational distribution adaptively as follows.

\section{METHODOLOGY}
\label{sec:method}
CATVI adopts the stochastic variational inference framework \citep{hoffman2013}, where the computation is accelerated by selecting a small batch of data and updating variational parameters with an unbiased random gradient. We first build the foundation of conditional variational inference as follows.

\subsection{Conditional Variational Inference}
\label{sec:CVD}
\paragraph{Conditional setting } HBNP models in (\ref{equ:model}) contain global latent variable $\beta$, local latent variables $\bz$, global prior $G_0$, local priors $\bG_{[J]}$ and observations $\bx$, where $\bz = \{\bz_j\}_{j=1}^J$, $\bz_j=\{z_{ji}\}_{i=1}^{N_j}$, $\bx = \{\bx_j\}_{j=1}^J$, $\bx_j=\{x_{ji}\}_{i=1}^{N_j}$ and $\bG_{[J]} =\{G_j\}_{j=1}^J$. We aim to find the variational distribution to maximize the NPELBO. 
In contrast to traditional approaches under the mean-field setting, we factorize the variational distribution as
\begin{equation}
\label{equ:condi_set}
\begin{gathered}
q(\beta, \bz, G_0, \bG_{[J]}) = q(\beta)q( G_0)\prod_{j=1}^{J}q(G_j | G_0, \bz_j)\prod_{i=1}^{N_j}q(z_{ji}),
\end{gathered}
\end{equation}
in the sense of the probability law. Such conditional design facilitates the recovery of the dependence structure among $G_0$, $\bG_{[J]}$ and $\bz.$ 

Combing (\ref{eq:ELBO}) and (\ref{equ:condi_set}), we seek to maximize the following NPELBO:
\begin{equation}
\label{equ:new_5}
    \begin{gathered}
      \liminf_{\vO}
    \Big\{  \E_{q(\beta, \bz, G_0^\vO, \bG_{[J]}^\vO)} \log  {p( \bx, \beta ,\bz, G_0^\vO, \bG_{[J]}^\vO)} 
    \\ 
    - \sum_{j=1}^{J}
    \E_{q(G_0^\vO)}\E_{q(\bz_j)} \E_{q(G_j^\vO | G_0^\vO, \bz_j)}   \log q(G_j^\vO | G_0^\vO, \bz_j)
    \\- \entropy \big(q(G_0^\vO)\big)
    - \entropy \big(q(\beta)\big)
    -  \sum_{j=1}^{J} \sum_{i=1}^{N_j} \entropy \big( q(z_{ji} )\big)\Big\},
    \end{gathered}
\end{equation}
where the entropy $\entropy\big(q(\cdot)\big) = E_{q(\cdot)} \log q(\cdot)$ and $\varOmega$ is a partition of the sample space $\Omega$ for $G_0$ and $\bG_{[J]}.$

\paragraph{Conditional variational distribution} To maximize the NPELBO in (\ref{equ:new_5}), we first compute the optimal variational distribution of $G_j$ given $G_0$ and $\bz_j$ for each $j$. 
As $p( \bx, \beta ,\bz, G_0^\vO, \bG_{[J]}^\vO)=p(G_0^\vO, \bz)p( \bx | \bz)\prod_{j=1}^{J}p(G_j^\vO | G_0^\vO, \bz_j)$, 
the non-constant term in (\ref{equ:new_5}) with respect to $q(G_j | G_0, \bz_j)$ is
\begin{equation*}
\begin{gathered}
    \liminf_{\vO} \Big\{
    \sum_{j=1}^{J} \E_{q(G_0^\vO)}\E_{q(\bz_j)} \E_{q(G_j^\vO | G_0^\vO, \bz_j)} \log   p(G_j^\vO | G_0^\vO, \bz_j)  \\
    - \log q(G_j^\vO | G_0^\vO, \bz_j)  \Big\}.
\end{gathered}
\end{equation*}
Note that the above expression can be viewed as the negative of a KL divergence whose maximum is zero. Therefore, to enable the NPELBO to reach the maximum, the optimal conditional variational distribution for $G_j$ should be $p(G_j | G_0, \bz_j).$ Consequently, NPELBO in (\ref{equ:new_5}) does not contain any term related to $q(G_j | G_0, \bz_j)$. 
In Appendix~\ref{app:1}, we derive NPELBO with respect to $q(G_0^\vO)$ as
\begin{equation}
\begin{gathered}
\label{eq:Conditional}
 \liminf_{\vO} \Big\{\sum_{j=1}^{J} \E_{q(G_0^\vO)} \E_{q(\bz_j)}  \log  \E_{p(G_j^\vO | G_0^\vO)}  p(\bz_j | G_j^\vO) \\
-\KL \big(q(G_0^\vO) \| p(G_0^\vO)\big) \Big\}
\end{gathered}
\end{equation}
up to a constant. It is important to note that $\E_{p(G_j^\vO | G_0^\vO)}$ is with respect to the prior $p(G_j^\vO | G_0^\vO)$ instead of the variational distribution $q(G_j^\vO | G_0^\vO)$, and hence this expectation can often be calculated analytically in HBNP models due to the conjugacy.

\subsection{Empirical Distribution and Evidence Lower Bound}
Within the conditional variational freamework, for the task of adaptive truncation, CATVI integrates Monte Carlo sampling to variational inference by  iterating the following steps till convergence, (i) using Monte Carlo sampling to get an empirical optimal variational distribution for local variables $\bz$ and (ii) updating the variational distributions for global variables $G_0$ and $\beta$.

\paragraph{Empirical distribution} From the entire data $\bx$, we randomly sample a subset $\{ \bx_s : \bx_s \in \bx \}_{s=1}^S$, where $S$ is the batch size with $S\ll J$. Given a partition $\varOmega$ in the current training iteration, we aim to update the parameters for $q(G_0^\vO)$ conditional on $q(\beta)$ and $\{q(\bz_s)\}_{s=1}^{S}$.
While standard stochastic variational inference updates parameters analytically, we use Monte Carlo sampling to draw $T_s$ samples for each $\bz_s$ from $q(\bz_s),$ 
thus constructing an empirical distribution,
\begin{equation*}
   \hat{q}(\bz_s) = \frac{1}{T_s}\sum_{t=1}^{T_s} \delta_{\hat{\bz}_{s,t}}, ~~ \hat{\bz}_{s,t} \sim q(\bz_s).
\end{equation*}

\paragraph{Empirical evidence lower bound} Using the empirical distribution $\hat{q}(\bz_s)$, we obtain an empirical evidence lower bound with respect to $q(G_0^\vO)$, $\widehat{\text{ELBO}}^{\vO},$ by replacing $q(\bz_s)$ in (\ref{eq:Conditional}) with $\hat{q}(\bz_s)$, that is,
\begin{equation}
\label{eq:stochastic_ELBO}
\begin{gathered}
\sum_{s=1}^{S}\sum_{t=1}^{T_s} \frac{J}{ST_s} \E_{q(G_0^\vO)}\log  \E_{p(G_s^\vO | G_0^\vO)}   p(\hat{\bz}_{s, t} | G_s^\vO)      \\ -\KL \big(q({G_0^\vO}) \| p({G_0^\vO})\big)
\end{gathered}
\end{equation} 
up to a constant. It is obvious that $\E(\widehat{\text{ELBO}}^{\vO})= \text{ELBO}^\vO,$ thus satisfying the key condition for stochastic variational inference \citep{hoffman2013}, that is, the random gradient is unbiased. 
Therefore, according to (\ref{eq:stochastic_ELBO}), we can use the random gradient generated from $\hat \bz_s =\{\hat{\bz}_{s,t}\}_{t=1}^{T_s}$ to update the parameters for $q(G_0^\vO)$. 

\paragraph{Resampling} We next present the procedure to get the empirical distribution $\hat{q}(\bz_s)$. As $G_s$ is integrated out, the local latent variables $\{z_{si}\}_{i=1}^{N_s}$ can not be sampled independently when we use Monte Carlo sampling to draw $\hat{\bz}_{s}$ given $q(G_0^\vO)$ and $q(\beta).$ Therefore, we propose the following Gibbs sampling approach to get samples under optimal variational distributions.
Conditional on $q(G_0^\vO)$, $q(\beta)$ and samples $\hat{\bz}_{s,i^-} = \{\hat{z}_{sl} : l=1,\dots, N_s, l \neq i\}$, 
it follows from (\ref{equ:new_5}) that the optimal variational distribution of $\log q(z_{si})$ is proportional to
\begin{equation}
\label{equ:gibbs_conditional}
\begin{gathered}
\E_{q({G_0^\vO})} \E_{p({G_j^\vO | G_0^\vO})}  p(z_{si}, \hat{\bz}_{s,i^-} | G_s^\vO)  \\ 
+ \E_{q({\beta})}  \log p(x_{si} | z_{si},\beta). 
\end{gathered}
\end{equation}
Then we sample $\hat{z}_{si} \sim q(z_{si})$ for each $i$ iteratively, which constructs a Markov chain. Noting that $q(z_{si})$ is often multinomial, sampling from its logarithm is commonly used. After the convergence, we can resample $\hat{\bz}_{s,1},\dots, \hat{\bz}_{s,T_s}$ from the stable Markov chain to update the parameters of $q(G_0^\vO)$ according to (\ref{eq:stochastic_ELBO}). Similarly, we derive the empirical evidence lower bound with respect to $q(\beta)$ in Appendix~\ref{app:catvi_algos} and can update the parameters for $q(\beta)$ using $\hat{\bz}_{s,1},\dots, \hat{\bz}_{s,T_s}$ correspondingly.

\subsection{Adaptive Truncation}
\label{sec:partition_re}
Finally, we seek to obtain the finite partition $\varOmega$ that could reach the limit inferior in NPELBO. Rather than having $\varOmega$ fixed on a universal truncation level, we enable the dimension of $\varOmega$ to gradually adjust to a stable level. This partition or truncation is dependent on data-fitting and embedded within the optimization process, providing another key advantage of using a Monte Carlo sampling scheme in the stochastic variational inference framework.

\paragraph{Partition refinement} According to the structure of HBNP models, $\hat{z}_{si}$ are sampled from the atomic support of $G_0$, $\{\phi_k \}_{k=1}^\infty$. Without loss of generality, we assume that the current partition $\varOmega$ consists of atomic elements $\phi_1, \dots, \phi_K \in \{\phi_k \}_{k=1}^\infty$ and their complement $\phi_0 = \Omega/ \{\phi_1, \dots, \phi_K \}$. Under this partition, $q(z_{si} \in \phi_0)$ is positive given (\ref{equ:gibbs_conditional}), and hence $\hat{z}_{si}$ can be sampled within $\phi_0$, that is, $\hat{z}_{si}$ is a new sample, distinct from  $\phi_1,\dots, \phi_K$. If this happens, we draw a new $\phi_{K+1}$ and refine the partition as $ \big(\phi_0 , \phi_1, \dots, \phi_K, \phi_{K+1}\big)$, where $\phi_0$ is updated as $\Omega/ \{\phi_1, \dots, \phi_K, \phi_{K+1} \}$.

\paragraph{Remark:} The partition refinement procedure reaches the limit inferior of empirical evidence lower bound as follows. Since there is no sampling within $\phi_0$ after each update, to minimize the KL divergence, the posterior should be proportional to the prior on $\phi_0$, $q\big(G_0(\phi_0)\big) \propto p\big(G_0(\phi_0)\big).$ 
Moreover, if we further partition $\phi_0$ into $\phi_0^{1} \cup \phi_0^{2}$, the KL divergence stays the same. Thus,  $\E(\widehat{\text{ELBO}}^{\vO})=\text{ELBO}^{\vO} = \text{NPELBO}.$  
See Appendix~\ref{app: justification} for a justification.

We summarize the CATVI algorithm in Algorithm~\ref{alg:1}.

\subfile{pictures/algo_table}

\section{APPLICATIONS IN TOPIC MODELS}
\label{sec:app_in_NLP}
\subsection{CATVI for the HDP Model}
\label{sec:infer_hdp}
We apply the proposed CATVI method to the HDP model. Specifically, we factorize the variational distributions in the conditional setting and specify the variational family as follows. First, the variational distribution of $G_s$ for each $s$ is given by $q(G_s | G_0, \bz_s)= \DirPro \big( \sum_{k=1}^{\infty}n_{sk}\delta_{\phi_k}  + G_0 \big) $, where $n_{sk} = \sum_{i=1}^{N_s} I( z_{si} = \phi_k)$ and $I(\cdot)$ is the indicator function. Second, $q(\beta_k)$ for each topic $k$ is set as a $W$-dimensional Dirichlet distribution, $q(\beta_k) = \text{Dirichlet}(\lambda_k)$, where 
$\lambda_k = (\lambda_{k1},\dots, \lambda_{kW})^{\T}$ is the parameter of vocabulary distribution for topic $k$. 
The variational distribution for topics without any observation remains the same as the prior, hence $q(\beta_0) = \text{Dirichlet}(\eta)$.
Third, we specify the variational family for $G_0$ using spike and slab distributions \citep{andersen2017} as
$
q(G_0) =  \sum_{k=1}^{K} m_k\delta_{\phi_k} + m_0 \DirPro (\alpha H ),
$
such that $\sum_{k=0}^K m_k =1$. 
Finally, following (\ref{equ:gibbs_conditional}) we use Monte Carlo sampling to obtain samples $\{\hat{\bz}_s\}_{s=1}^S$, avoiding the need to parametrize their variational distributions. 

As different samples in $\{\hat{\bz}_s\}_{s=1}^S$ are used to represent different topic clusters in topic modelling, their exact values in the sample space do not contain any statistical information. We can then simply index the topics from $1$ to $K$ and denote the different clusters by distinct points $\phi_1, \dots, \phi_K$ in $\Omega,$ and cluster $0$ is the topic without any observation.
Given samples $\{\hat{\bz}_s\}_{s=1}^S$, we define the number of topics with observations by $K=\sum_{k=0}^{\infty} I(\sum_{s=1}^{S} \sum_{t=1}^{T_s} \hat{n}_{sk,t} > 0)$, where $\hat{n}_{sk, t} = \sum_{i=1}^{N_s} I (\hat{z}_{si,t} = \phi_k).$
In Appendix~\ref{app:2}, we rely on (\ref{eq:stochastic_ELBO}) to derive the empirical evidence lower bound with respect to $q(G_0),$ 
\begin{equation}
\label{equ:hdp_NPELBO}
\begin{gathered}
\alpha \log m_0 - \sum_{k=0}^{K}  \log m_k  \\ + 
\sum_{s=1}^{S} \sum_{k=1}^{K} \sum_{t=1}^{T_s}\frac{J}{ST_s} \log  \frac{ \Gamma(\gamma m_k + {\hat{n}_{sk,t}})}{\Gamma(\gamma m_k) }
\end{gathered} 
\end{equation} 
up to a constant. 
According to Algorithm~\ref{alg:1}, we repeatedly select documents of a batch size $S$, sample $\{\hat{\bz}_s\}_{s=1}^S$, and update parameters for $G_0$ and $\beta$ iteratively until the empirical evidence lower bound converges to its maximum. During Gibbs sampling, once a document is sampled in cluster $0$, we add a new cluster $K+1,$ thus partitioning $\varOmega$ to be $(K+1)$-dimensional, with $K$ single points $\{\phi_k\}_{k=1}^K$ and one complement set $\phi_0 = \Omega/ \{\phi_k\}_{k=1}^K.$ During the training, this procedure is repeated until $\varOmega$ is optimized.
See Appendix~\ref{app:catvi_for_hdp} for detailed steps on updating the variational parameters and refining the partition.

\subsection{CATVI for Generic HBNP Models}
\label{sec:extensions_CRM}
The CATVI algorithm can also be applied to a general class of HBNP models, where the global prior $G_0$ is generated from a completely random measure. In these models, the concentration parameter for any $G_j$ is not fixed, and $G_0$ is not restricted to be a probability measure. The corresponding inference algorithm is similar to that of the HDP model, but requires a new parameter $\mu$ to approximate $G_0(\Omega)$. We choose the variational family for the global prior $G_0$ as 
$
q(G_0) =\mu \big(\sum_{k=1}^{K} m_k\delta_{\phi_k} + m_0 \widetilde{N}(\alpha H )\big),
$
where $\widetilde{N}$ is the normalization of the corresponding completely random measure and $\sum_{k=0}^{K}m_k = 1$. 
We provide the corresponding empirical evidence lower bounds and algorithms to infer more general HBNP models, 
including $\Gamma$DP model, in Appendix~\ref{alg:GammaDP-algo}.

\section{RELATIONSHIP TO RELATED WORKS}
\label{sec:comparision}
In this section, we discuss several advantages of 
CATVI compared with traditional methods  \citep{hoffman2013, wang2011, wang2012}, although these are specific to the inference for the HDP model. First, CATVI replaces the unrealistic mean-field assumption with the conditional setting to capture the correlation structure among latent variables. 
Second, CATVI approximates the posterior groupwisely instead of updating the stick-breaking parameters sequentially, and hence avoids the gradient vanishing problem. By contrast, \citet{hoffman2013} and \citet{wang2011} perform inference separately over each atomic location and weight of $G_0$ using the stick-breaking representation $G_{0K} = g_{0K}\prod_{k=1}^{K-1}(1-g_{0k})$, where $g_{0k}$s are the representation parameters. However, this may cause the gradient vanishing problem of $G_{0K}$ if $k$ is large, because $\prod_{k=1}^{K-1}(1-g_{0k})$ is close to zero. 
Third, these traditional methods universally truncate the dimension of $G_0$ to a fixed level, contradicting the motivation and advantages of using HBNP models. Finally, CATVI is guaranteed to maximize the NPELBO. By comparison, \citet{wang2012} update parameters using the locally collapsed Gibbs sampling, but their work leads to an approximation that fails to maximize the ELBO, especially when the variance of distributions is large. 

From a computational perspective, CATVI inherits the fast speed of stochastic variational inference, while other methods that truncate the dimension in a truly nonparametric way are very slow, such as the split-merge variational inference \citep{bryant2012} and the pure Gibbs sampling \citep{teh2006}. 
To check the split-merge criterion, the split-merge variational inference requires calculating the likelihood before and after a split or merge, which is computationally infeasible in practice. Moreover, the pure Gibbs sampling is not scalable as well. As pure Gibbs sampling does not have batch selection, the Markov chains would converge very slowly when the sample size is large. As a result, these methods cannot be used to handle big data.

\section{EXPERIMENTS}
\label{sec:real_data_analysis}
\subsection{Datasets and Architectures}
We apply the CATVI algorithm to three large datasets, {\textit{arXiv}}, {\textit{NYT}} and {\textit{Wiki}}, and compare the performance of CATVI with the online variational inference (OVI) \citep{wang2011}, the memorized online variational inference (MOVI) \citep{hughes_memoized_2013},  the split-merge variational inference (SMVI) \citep{bryant2012} and Gibbs sampling (GS) \citep{teh2004}. 
\paragraph{\textit{arXiv}} The corpus contains descriptive metadata of articles on \textit{arXiv} 
up to September 1, 2019, resulting in 1.03M documents and 44M words from a vocabulary of 7,500 terms.
\paragraph{\textit{NYT}} The corpus contains all articles published by \textit{New York Times} from January  1987 to June 2007 \citep{sandhaus2008}, resulting in 1.56M documents and 176M words from a vocabulary of 7,600 terms.
\paragraph{\textit{Wiki}} The corpus contains entries from all English \textit{Wikipedia} websites on January 1, 2019, resulting in 4.03M documents and 423M words from a vocabulary of 8,000 terms. 

For the preprocessing, stemming and lemmatization are used to clean the raw text, and then words with too high or too low frequency, as well as common stop words, are filtered out.

\begin{figure}[!ht]
\begin{subfigure}{1.0\linewidth}
\caption{\textit{arXiv}}
\label{fig:result_wiki}
\includestandalone[width=0.48\linewidth]{plot_1}
\includestandalone[width=0.48\linewidth]{plot_4}
\end{subfigure}
\begin{subfigure}{1.0\linewidth}
\caption{\textit{NYT}}
\label{fig:result_arxiv}
\includestandalone[width=0.48\linewidth]{plot_2}
\includestandalone[width=0.48\linewidth]{plot_5}
\end{subfigure}
\begin{subfigure}{1.0\linewidth}
\caption{\textit{Wiki}}
\label{fig:result_nyt}
\includestandalone[width=0.48\linewidth]{plot_3}
\includestandalone[width=0.48\linewidth]{plot_6}
\end{subfigure}
\caption{Left column: plots for the perplexity vs the running time up to 5 hours, Right column: plots for the number of topics vs the running time. 
}
\label{fig:result_0}
\end{figure}
To evaluate the performance of CATVI, we set aside a test set of 10,000 documents for each dataset and calculate the predictive perplexity as
$$
\text{perplexity} = 
\exp\Big\{{ -\frac{\sum_{j\in \text{D}_{\text{test}}}
\log p(\bx_{j}^{\text{test}} | \bx_{j}^{\text{train}}, \text{D}_{\text{train}} )
}{\sum_{j\in \text{D}_{\text{test}}} |\bx_{j}^{\text{test}}|}
}\Big\},
$$
where $\text{D}_{\text{train}}$ and $\text{D}_{\text{test}}$ represent the training and test data, respectively, $\bx_{j}^{\text{train}}$ and $\bx_{j}^{\text{test}}$ are the training and test words in test document $j,$ respectively, and $|\bx_{j}^{\text{test}}|$ is the number of words in $\bx_{j}^{\text{test}}$ \citep{ranganath2018}. 
The perplexity measures the uncertainty of fitted models, where a lower perplexity will result in a better language model with higher predictive likelihood.
Since the perplexity can not be computed exactly, the standard routine uses $\text{D}_{\text{train}}$ to compute the variational distribution for $\beta$ and $G_0$, then obtains the variational distribution for $G_j$ based on $G_0$ and $\bx_{j}^{\text{test}}$, and then approximates the likelihood by $p(\bx_{j}^{\text{test}} | \bx_{j}^{\text{train}}) = \prod_{w \in \bx_{j}^{\text{test}}}\sum_{k=0}^{K} \overline{G}_{jk} \overline{\beta}_{kw}$, where $\overline{G}_{jk}$ and $\overline{\beta}_{kw}$ are the variational expectations of ${G}_{jk}$ and $\beta_{kw}$, respectively \citep{blei2003}.
Experiments are run with the three datasets above using both the HDP and $\Gamma$DP models. For the HDP model, we set the hyperparameters as $\alpha=\gamma=\eta=5$, where $\alpha$ and $\gamma$ are the concentration parameters for $G_0$ and $G_j$ respectively, and $\eta$ is the hyperparameter for the prior on the distribution of words. The initial number of topics is set to be 100. The parameters are then optimized using stochastic gradient descent, with a batch size of 256 and a linear decaying  learning rate adopted in \citet{hoffman2010}. 
For the $\Gamma$DP model, we use the same settings but discard $\gamma$. 
In the experiments, we remove clusters with fewer than 1 document during the training.

\subsection{Empirical Results}
\paragraph{Predictive perplexity} The top row of Figure~\ref{fig:result_0} plots the predictive perplexity as a function of running time for the three  comparison methods using the three datasets. As MOVI,  SMVI and GS provide much higher perplexities, we do not plot their results in Figure~\ref{fig:result_0}. Table~\ref{tab:result} reports numerical summaries for all comparison methods. In particular, as GS can not scale to large datasets, we use a subset with 500 documents to run the experiments.
Several conclusions can be drawn here. 
First, on all three datasets, CATVI uniformly outperforms competing methods. 
The improvement is highly consequential, especially for \textit{arXiv} and \textit{Wiki}. For \textit{NYT}, there is moderate improvement, likely due to the long length of documents in this corpus. 
Second, for each dataset, the $\Gamma$DP model attains a lower perplexity than the HDP model, consistent with the fact that the $\Gamma$DP model removes a restriction of the HDP model and hence is more flexible.
Third, CATVI is empirically shown to be computationally efficient, reaching the lowest perplexity within the same training time.  Although it involves Monte Carlo sampling, the perplexity converges fast. This is because the convergence of local Markov chains to assign words to topics is accelerated by a clear topic-words clustering as the global variational distributions approach to the optimal. 
\begin{table}[t]
\caption{A summary of predictive perplexity results. 
}
\label{tab:result}
\begin{center}
\begin{small}
\begin{sc}
\begin{tabular}{lcccr}
\toprule
 Model & Method &  \textit{arXiv}&  \textit{NYT}  & \textit{Wiki} \\ 
\midrule
HDP & GS & 3175 & 2635 &  1807\\
HDP & MOVI & 1901 & 2921 & 1876 \\
HDP & SMVI & 1917 & 2866 & 1877 \\
HDP & OVI & 1005 & 1681 & 1422 \\ 
HDP & CATVI & 832 & 1569 & 1207 \\
$\Gamma$DP & CATVI & 808 & 1536 & 1157 \\
\bottomrule
\end{tabular}
\end{sc}
\end{small}
\end{center}
\vspace{-0.1in}
\end{table}

\paragraph{Number of topics} The bottom row of Figure~\ref{fig:result_0}  plots the number of topics during the training process. For OVI, the number of topics remains constant at the prespecified value, while for CATVI, this value first increases steeply and then converges to a stable level. For example, the number of topics in \textit{Wiki} sharply increase from 100 to around 190 for the HDP model and around 200 for $\Gamma$DP model. 
The sharp increase is driven by the data complexity, while the stable level is achieved due to the dimension penalty effect from the priors in HBNP models.
Although the estimation of the number of topics is not consistent, CATVI can provide some useful information about topics in data. For instance, the data from the \textit{arXiv} corpus in these experiments are limited to abstracts of scientific articles, and thus it has the smallest number of topics. 
By contrast, \textit{NYT} is a compilation of all new articles covering a wider range of areas, and hence consists of more topics. Similarly, \textit{Wiki} has the largest number of topics as it contains almost every aspect of an encyclopedia. 
It is important to note that we do not need to set a fixed number of topics before the inference. Instead, CATVI starts from an initial value, for example 100 in our experiments, then automatically converges to a stable optimal number of topics.

\paragraph{Topic-words clustering} CATVI is shown to reveal much better linguistic results. To compare CATVI with OVI for the HDP model, we report the top 12 words in the top 10 topics with biggest weights 
for both methods on \textit{arXiv} and \textit{Wiki} 
in Tables~\ref{tab:result_topics_arxiv}~and~\ref{tab:result_topics_wiki}, respectively. We observe a few apparent patterns.
First, the topic-word clusters from CATVI hardly contains replicated topics, whereas those from OVI results have similar word components, such as those shown in blue in columns 1-6 in the bottom part of Table~\ref{tab:result_topics_arxiv}. 
An ideal topic-word clustering should allocate these words into just one topic. However, the prespecified number of topics is fixed at 150 in OVI, which is larger than the ground truth, resulting in  generating replicated topics. By contrast, the topic-word clustering by CATVI does not have such redundancy. It is apparent that our top 10 topics are mostly distinct.
Second, CATVI leads to much clearer topic-word clustering. For both datasets, our results indicate that all of our detected words within any column are highly relevant and should intuitively be grouped into one cluster with clear linguistic meaning. For example, column 7 of Table~\ref{tab:result_topics_wiki} for CATVI presents several words all related to military, but words in the same column for OVI seem to be a mixture of several loosely connected topics including `human, character, reveal, episode, comic, voice', `human, earth' and `human, kill, attack, fight, battle, doctor'. This mixture of topics makes the topic-word clustering in this column ambiguous. 
Furthermore, CATVI identifies a topic about popular English given names in column 5 of Table~\ref{tab:result_topics_wiki}. Although these given names are not shown in a single document, CATVI can successfully discover that they belong to one topic, while OVI fails.
This is because CATVI does not force the topics to merge together if the prespecified number of topics is not large enough, thus reducing the noise in the clusters.


We also perform sensitivity analysis of CATVI using \textit{arXiv} under the HDP model as an example. The left and right panels of Figure~\ref{fig:hpo} in Appendix~\ref{app: sensitivity} respectively plot the results as the batch size varies from 128 to 1024 and the initial number of topics varies from 60 to 140. We observe that the performance is not sensitive to the change of these hyperparameters. Moreover, the best results are obtained for the case with a smaller batch size and a larger initial number of topics.

\section{DISCUSSION}
\label{sec.dis}
CATVI can also be applied to other HBNP models including, for example, hierarchical Pitman--Yor process model \citep{teh2010} and hierarchical beta process model \citep{thibaux2007}. 
CATVI will provide more advantages in these applications,
because the hierarchical Pitman--Yor process, with heavy tail behavior, and the hierarchical beta process, with sparse structure, may suffer more from the universal truncation.

\begin{table}
\caption{Top 12 words in top 10 topics.}
\label{tab:result_topics}
\centering
\begin{subtable}{0.99\linewidth}
\caption{\textit{arXiv}}
\centering
\label{tab:result_topics_arxiv}
    \includestandalone[width=\linewidth]{Topics_arXiv}

\end{subtable}
\vskip 0.05in
\begin{subtable}{0.99\linewidth}
\caption{\textit{Wiki} }
\centering
\label{tab:result_topics_wiki}
    \includestandalone[width=1.0\textwidth]{Topics_Wiki}

\end{subtable}
\end{table}

\subsubsection*{Acknowledgments}
We thank the anonymous reviewers
for useful comments during the review process.

Opinions expressed in this paper are those of the authors, and do not necessarily reflect the view of J.P. Morgan.
Opinions and estimates constitute our judgement as of the date of this Material, are for informational
purposes only and are subject to change without notice. This Material is not the product of J.P.
Morgan’s Research Department and therefore, has not been prepared in accordance with legal
requirements to promote the independence of research, including but not limited to, the prohibition
on the dealing ahead of the dissemination of investment research. This Material is not intended as
research, a recommendation, advice, offer or solicitation for the purchase or sale of any financial
product or service, or to be used in any way for evaluating the merits of participating in any transaction.
It is not a research report and is not intended as such. Past performance is not indicative of future
results. Please consult your own advisors regarding legal, tax, accounting or any other aspects
including suitability implications for your particular circumstances. J.P. Morgan disclaims any
responsibility or liability whatsoever for the quality, accuracy or completeness of the information
herein, and for any reliance on, or use of this material in any way.
Important disclosures at: www.jpmorgan.com/disclosures.

\bibliography{document.bib}


\clearpage
\appendix

\thispagestyle{empty}

\onecolumn \makesupplementtitle

\appendix
This supplementary material contains a short review of completely random measures in Appendix~\ref{app:CRM}, CATVI algorithm and its applications to the HDP model and the $\Gamma$DP model in Appendix~\ref{app:catvi algorithms}, technical proofs and derivations in Appendix~\ref{ap_proof}, computational complexity analysis and code in Appendix~\ref{ap_comp} and sensitivity analysis results in Appendix~\ref{app: sensitivity}.

\section{A Short Review of Completely Random Measures}
\label{app:CRM}
\setcounter{equation}{0}
\renewcommand\theequation{A.\arabic{equation}}
Suppose that $(\Omega,\cF)$ is a Polish sample space, $\Theta$ is the set of all bounded measures on $(\Omega,\cF)$ and $\cM$ is a $\sigma$-algebra on $\Theta$. A random measure $G$ on $(\Omega,\cF)$ is a transition kernel from $(\Theta, \cM)$ into $(\Omega,\cF)$ such that (i) $G \mapsto G(A)$ is $\cM$-measurable for any $A \in \cF$ and (ii) $A \mapsto G(A)$ is a measure for any realization of $G$ \citep{ghosal2017}. For example, a Dirichlet process $P$ 
with base measure $P_0$ satisfies
\begin{equation*}
    \big(P(A_1),\dots ,P(A_n)\big) \sim \text{Dirichlet} \big( P_0(A_1),\dots ,P_0(A_n)\big)
\end{equation*}
for any partition $\varOmega = (A_1,\dots,A_n)$ of $\Omega$, that is, a finite number of measurable, nonempty and disjoint sets such that $\bigcup _{i=1}^{n}A_i = \Omega$. The Dirichlet process is denoted by $P \sim \DirPro(P_0)$ or $P \sim \DirPro(\alpha H)$ with concentration parameter $\alpha = P_0(\Omega)$ and center measure $H = \alpha^{-1}P_0$. Moreover, a random measure is called a completely random measure \citep{Kingman1993} if it also satisfies the condition that (iii) $P(A_i)$ is independent of $P(A_j)$ for any disjoint subsets $A_i$ and $A_j$ in $\Omega$. Completely random measures and their normalizations \citep{ghosal2017}, for example, the Gamma process and Dirichlet process, respectively, are commonly used as priors for infinite-dimensional latent variables in HBNP models, because their realizations are atomic measures with countable-dimensional supports.

A completely random measure \citep{Kingman1993} is characterized by its Laplace transform,
\begin{equation*}
\E\big[e^{-tP(A)}\big] = \exp \Big\{- \int_{A} \int_{(0, \infty]} (1- e^{-t\pi})v^c (dx , ds) \Big\},
\end{equation*}
where $A$ is any measurable subset of $\Omega$ and $v^c(dx , ds)$ is called the L\'evy measure. If $v^c(dx , ds) = \kappa(dx) v(ds)$, where $\kappa(\cdot)$ and $v(\cdot)$ are measures on $\Omega$ and $(0, \infty]$, respectively, the completely random measure is homogeneous \citep{ghosal2017}. In such a case, we call $v(\cdot)$ the weight intensity measure.
We can view completely random measure as a Poisson process on the product space $\Omega \times (0, \infty]$ using its L\'evy measure as the mean measure.

\section{CATVI Algorithm}
\label{app:catvi algorithms}
\subsection{Empirical ELBO for $q(\beta)$ and $q(z_{si})$}
\label{app:catvi_algos}
To maximize the NPELBO, we iterate the following three steps: (i) randomly select a small batch from the entire data, (ii) sample $\{\hat{\bz}_s\}_{s=1}^{S}$ by Monte Carlo method, and (iii) update $q(G_0^\vO)$ and $q(\beta)$ in the stochastic variational inference framework.

In an analogy to (\ref{eq:Conditional}), the NPELBO with respect to $q(\beta)$ is
\begin{equation}
\label{equ:sample_beta_new}
\liminf_{\vO} \Big\{\sum_{j=1}^{J} \E_{q(\beta)} \E_{q(\bz_j)} \text{log} p(\bx_j| \bz_j, \beta)
-\KL\big(q(\beta) \| p(\beta)\big) \Big\},
\end{equation}
and the empirical evidence lower bound with respect to $q(\beta)$, $\widehat{\text{ELBO}}^{\vO}$, is,
\begin{equation}
\begin{gathered}
\label{eq:stochastic_beta_ELBO}
\sum_{s=1}^{S} \sum_{t=1}^{T_s} \frac{J}{ST_s}\E_{q(\beta)} \log p(\bx_s | \hat{\bz}_{s,t}, \beta) 
-\KL \big(q({\beta})|p({\beta})\big)
\end{gathered}
\end{equation}
up to a constant and then we can update its parameter with the corresponding random gradient in a similar way. Moreover,  the NPELBO with respect to $q(z_{si})$ is 
\begin{equation}
\liminf_{\vO} \Big\{ \E_{q(G_0^\vO)} \E_{q(\bz_j)}  \log  \E_{p(G_j^\vO | G_0^\vO)}  p(\bz_j | G_j^\vO) \\
+ \E_{q(\bz_j)} \E_{q(\beta)}  \log p(\bx_{j} | \bz_{j},\beta)  
- \E_{q(\bz_j)} \log q(\bz_j)
\Big\}.
\end{equation}
Factorizing $\E_{q(\bz_j)}$ as $\E_{q(\bz_{ji})}\E_{q(\bz_{j, i^-})}$ leads to (11). 
We summarize the details of CATVI algorithm in Algorithm~\ref{alg:1}.

\subsection{CATVI for the HDP Model}
We repeatedly select documents of a batch size and update parameters iteratively according to the following three steps, until the NPELBO attains its maximum.

\paragraph{Inference for $G_0.$} 
\label{app:catvi_for_hdp}
\setcounter{equation}{0}
\renewcommand\theequation{B.\arabic{equation}}
There is no closed-form expression for the parameters $\{m_k\}_{k=0}^K$ to attain the maximum in (\ref{equ:hdp_NPELBO}). Moreover, the standard gradient descent algorithm fails in this case, because  $\{m_k\}_{k=0}^K$ may easily exceed the simplex during the updating procedure. Instead, given the parameters $\{m_k^{(\tau)}\}_{k=0}^{K}$ in the $\tau$-th iteration, we first define
\begin{equation}
\label{equ:hdp_gradient}
m_{k}^{*} \propto  
\begin{cases}
    JS^{-1}\gamma \sum_{s=1}^{S} \big\{ T_s^{-1} \sum_{t=1}^{T_s}\Phi(\gamma m_k^{(\tau)} + {\hat{n}_{sk,t}})-\Phi(\gamma m_k^{(\tau)})\big\} m_k^{(\tau)}-1 & k=1, \dots, K,\\
    \alpha -1              & k = 0,
\end{cases}
\end{equation}
where $\Phi(\cdot)$ denotes the log-gamma function,
such that $\sum_{k=0}^{K}m_{k}^{*} =1$, and then we update the parameters by $m^{(\tau+1)} = (1-\rho_t) m^{(\tau)} + \rho_\tau m_{k}^{*},$ where $\rho_t$ is the step size defined in Algorithm \ref{alg:1}. This updating algorithm is consistent to the gradient descent after the inverse logit transformation. See Appendix \ref{app:3} for a justification. In the process of updating, the condition $\sum_{k=0}^{K}m_k^* = 1$ always holds, and hence we eliminate the risk of exceeding the simplex.

 \paragraph{Inference for $\beta.$} 
 By (\ref{eq:stochastic_beta_ELBO}), we update the parameters for $q(\beta)$ using samples $\{\hat{\bz}_s\}_{s=1}^S$. We define $\lambda_{kw}^*$ for topic $k$ and word $w$ as,
\begin{equation}
\label{eq:inference_for_beta_in_appendix}
	\lambda_{kw}^* = \eta + \sum_{s=1}^{S} \sum_{t=1}^{T_s} \sum_{i=1}^{N_s} \frac{J}{ST_S}  I( \hat{z}_{si,t} = \phi_k, x_{si}=w), 
\end{equation}
and update the parameter $\lambda_k$ by $\lambda_k^{(\tau+1)} =(1-\rho_t)  \lambda_k^{(\tau)}  + \rho_\tau \lambda_k^*$ for each $k$, where $\lambda_k^*=(\lambda_{k1}^*, \dots, \lambda_{kW}^*)^{\T}$.

 \paragraph{Sampling for $\bz.$} 
According to (\ref{equ:gibbs_conditional}) we sample $\hat{z}_{si}$ conditional on $q(G_0)$ and $\hat{\bz}_{si^{-}}$ by
\begin{equation}
\label{equ:sampleing_z-hdp}
q(z_{si}= \phi_k)  \ \propto \  
\begin{cases}
(\gamma m_k + \hat{n}_{s,i^{-}}^{k} ) \exp \big( \Phi(\lambda_{kx_{si}}) - \Phi(\sum_{w=1}^{W}\lambda_{kw})\big)  & k=1, \dots, K,\\
\gamma m_0  \exp \big( \Phi(\eta) - \Phi(W\eta)\big)  & k = 0,
\end{cases}
\end{equation}
to construct the Markov chain, where $\hat{\bn}_{s,i^{-}}^{k}=\sum_{1 \leq l \leq N_s, l \neq i}I(\hat{z}_{sl}=\phi_k)$. Whenever the sampled $\hat{z}_{si}$ is in $\phi_0$, meaning $\hat{z}_{si}$ forms a new point not belonging to $\{\phi_1, \dots, \phi_K\}$, we need to update the partition and add a new topic indicated by $\phi_{K+1}$. Otherwise the partition dimension remains the same. Iterating the sampling scheme till convergence, we obtain the samples $\{\hat{z}_{si, t}\}_{1\leq s \leq S, 1\leq i \leq N_s, 1\leq t\leq T_s}$ and corresponding $\{\hat{n}_{sk, t}\}_{1\leq s \leq S, 1\leq k \leq K, 1\leq t\leq T_s}$ for the selected chunk.

\subsection{CATVI for the $\Gamma$DP Model}
\label{alg:GammaDP-algo}

$\Gamma$DP releases the constraint of fixed concentration parameter $\gamma$ in HDP. 
Therefore, 
the CATVI algorithm for 
$\Gamma$DP inherits the steps in (\ref{equ:hdp_gradient}) and (\ref{equ:sampleing_z-hdp}), 
except that a parameter $\mu$ replaces the concentration parameter $\gamma$ in both formulas. 

We derive the empirical evidence lower bound in Appendix~\ref{app:extensions} with respect to $q(G_0)$ as,
\begin{equation}
\begin{gathered}
\label{equ:general_case}
\sum_{k=1}^{K}
\log v(\mu m_k) 
+ \log u(\mu m_0) +  
\sum_{s=1}^{S}\frac{J}{S} \log \frac{{\Gamma}(\mu)}{{\Gamma}(\mu+N_s)}  
  + \sum_{s=1}^{S}   \sum_{k=1}^{K} \sum_{t=1}^{T_s}  \frac{J}{ST_s} \log \frac{ {\Gamma}( \mu m_k + {\hat{n}_{sk,t}})}{\Gamma(\mu  m_k) }  + K\log \mu 
\end{gathered}
\end{equation} 
up to a constant, where $v(\cdot)$ is the weight intensity measure (see Appendix~\ref{app:CRM}) for the completely random measure, and $u(\cdot)$ is the density function for $G_0(\Omega)$ that can be derived using its Laplace transform. Therefore, we can update $\{m_k\}_{k=0}^{K}$ in the same way as the HDP model. 

Similar to (\ref{equ:hdp_gradient}), we update $\{m_k\}_{k=0}^K$ according to 
\begin{equation}
\label{equ:gamma_dp_gradient}
m_{k}^{*} \propto  
\begin{cases}
    JS^{-1}\mu \sum_{s=1}^{S} \big\{ T_s^{-1} \sum_{t=1}^{T_s}\Phi(\mu m_k^{(\tau)} + {\hat{n}_{sk,t}})-\Phi(\mu m_k^{(\tau)})\big\} m_k^{(\tau)}-1 & k=1, \dots, K,\\
    \alpha -1              & k = 0,
\end{cases}
\end{equation}
and $m^{(\tau+1)} = (1-\rho_t) m^{(\tau)} + \rho_\tau m_{k}^{*}.$ Moreover, in an analogy to (\ref{equ:sampleing_z-hdp}), the probability to sample $\hat{z}_{si}$ is defined as
\begin{equation}
\label{equ:gamma_sampleing_z-hdp}
q(z_{si}= \phi_k)  \ \propto \  
\begin{cases}
(\mu m_k + \hat{n}_{s,i^{-}}^{k} ) \exp \big( \Phi(\lambda_{kx_{si}}) - \Phi(\sum_{w=1}^{W}\lambda_{kw})\big)  & k=1, \dots, K,\\
\mu m_0  \exp \big( \Phi(\eta) - \Phi(W\eta)\big)  & k = 0.
\end{cases}
\end{equation}
Finally, we apply the gradient ascent to update $\mu$. In Appendix~\ref{app:extensions}, we derive the gradient of empirical evidence lower bound with respect to $\mu$ as 
\begin{equation}
\label{eq:gamma_mu_update}
\begin{split}
g'(\mu)= 
\frac{\alpha-1}{\mu} -1 + \frac{J}{S}\sum_{s=1}^{S}  \Big\{ \Phi({\mu})-\Phi({\mu+N_s}) +  \sum_{k=1}^{K} \frac{1}{T_s} \sum_{t=1}^{T_s} m_k\big(\Phi( \mu m_k + {\hat{n}_{sk,t}})-\Phi(\mu  m_k  ) \big)\Big\},
\end{split}
\end{equation}
and then update $\mu$ by $\mu^{(\tau+1)} = \mu^{(\tau)} + \rho_\tau g'(\mu^{(\tau)}).$

\section{Technical Proofs and Derivations}
\label{ap_proof}
\setcounter{equation}{0}
\renewcommand\theequation{C.\arabic{equation}}
\subsection{Proof for (\ref{equ:5})}
\label{app:induced}
By definition of induced measure, $q^\vO(d\Theta) = Q(d\Theta)$ for any $\cM$-measurable $d\Theta$ , we have
\begin{eqnarray*}
  \int_{\Theta}  \log \frac{dq^\vO}{dp^\vO}  dq^\vO =  \int_{\Theta}   \log \frac{dq^\vO}{dp^\vO}   dQ.
\end{eqnarray*}
It follows from $\limsup_{\vO } dq^\vO / dp^\vO = dQ / dP$ and the monotone convergence theorem that
\begin{eqnarray*}
  \limsup_{\vO} \int_{\Theta}   \log \frac{dq^\vO}{dp^\vO}  dQ =  \int_{\Theta}  \log \frac{dQ}{dP} dQ.
\end{eqnarray*}
Combining the above equations yields (\ref{equ:5}). Furthermore, suppose there exists a sequence of partition $\{\varOmega_i\}_{i \geq 1}$ such that $\limsup \varOmega_i = \varOmega$, we have
\begin{eqnarray*}
\begin{aligned}
  \limsup_{\vO_i} \int_{\Theta}   \log \frac{dq^{\vO_i}}{dp^{\vO_i}}  dq^{\vO_i} =  
  \limsup_{\vO_i} \int_{\Theta}  \log \frac{dq^{\vO_i}}{dp^{\vO_i}}  dQ  =  \int_{\Theta}   \log \frac{dq^\vO}{dp^\vO}   dQ
  = \int_{\Theta}  \log \frac{dq^\vO}{dp^\vO}  dq^\vO.
 \end{aligned}
\end{eqnarray*}
Hence $\limsup_{\vO_i} \text{KL}(q^{\vO_i} \| p^{\vO_i}) = \text{KL}(q^{\vO} \parallel p^{\vO}),$ which will be used in Appendix~\ref{app:2}.

\subsection{Proof for (\ref{equ:7})}
\label{app:induced_EBLo}
By $p(X, Z)=   p( Z | X) p(X) $, we have 
\begin{eqnarray*}
    \int   \log \frac{p(X,Z^\vO)}{q(Z^\vO)}q(dZ ^\vO)  =\log p(X) +  \int   \log \frac{p(Z^\vO | X)}{q(Z^\vO)}q(dZ ^\vO).
\end{eqnarray*}
Taking the limit inferior on both sides, we have
\begin{eqnarray*}
\begin{aligned}
    \liminf_{\vO}   \int    \log \frac{p(X,Z^\vO)}{q(Z^\vO)}q(dZ ^\vO)   =\log p(X) - \limsup_{\vO} \bigg\{ -  \int \log \frac{p(Z^\vO | X)}{q(Z^\vO)}q(dZ ^\vO) \bigg\}.
\end{aligned}
\end{eqnarray*}
Combing the above equation with the definition of NPELBO in (\ref{eq:ELBO}) and  the KL divergence in (\ref{equ:5}) yields (\ref{equ:7}).

\subsection{Derivation for (\ref{eq:Conditional})}
\label{app:1}
By
$
p(G_0^\vO, \{\bz_j\}_{j=1}^{J}) = \int \cdots \int p\big(G_0^\vO, \{G_j \}_{j=1}^{J},\{\bz_j\}_{j=1}^{J}\big) dG_1 dG_2 \cdots dG_J
$ and the hierarchical generative structure, the evidence lower bound under partition $\varOmega$ with respect to $q(G_0^\vO)$ equals,
\begin{eqnarray*}
    & &\text{ELBO}^\vO \\
    &=&
        \E_{q(G_0^\vO)}\E_{q(\{\bz_j\}_{j=1}^{J})}\log p(G_0^\vO, \{\bz_j\}_{j=1}^{J})   - \E_{q(G_0^\vO)}  \log q(G_0^\vO) +\text{constant} \\
    &=&
        \E_{q(G_0^\vO)}\E_{q(\{\bz_j\}_{j=1}^{J})}\log  q(G_0^\vO) \prod_{j=1}^{J} \int p(G_j^\vO | G_0^\vO) p(\bz_j | G_j^\vO)dG_j  -  E_{q(G_0^\vO)}  \log q(G_0^\vO)  +\text{constant}  \\
    &=&
        \sum_{j=1}^{J} \E_{q(G_0^\vO)}\E_{q(\bz_j)}   \log  \E_{p(G_j^\vO | G_0^\vO)} p(\bz_j | G_j^\vO)   
       + \E_{q(G_0^\vO)} \log p(G_0^\vO) - \E_{q(G_0^\vO)} \log q(G_0^\vO) +\text{constant}. 
\end{eqnarray*}
Furthermore, based on the equation above, (\ref{equ:new_5}) can be expressed as $\text{NPELBO}= \liminf_{\vO} \text{ELBO}^{\vO}$.

\subsection{Derivation for (\ref{equ:hdp_NPELBO})}
\label{app:2}
By the formula of moments for Dirichlet-distributed random variables, we obtain
\begin{equation*}
    \E_{p(G_s^\vO | G_0^\vO)} p(\hat{\bz}_{s,t} | G_s^\vO)  = 
     \frac{\Gamma(\gamma )}{\Gamma(\gamma +N_s)} \prod_{k=1}^{K} \frac{\Gamma(\gamma G_{0k}  +\hat{n}_{sk,t})}{\Gamma(\gamma G_{0k})}.
\end{equation*}
Based on the points $\{\phi_k \}_{k=1}^K$ defined in Section \ref{sec:infer_hdp}, we propose a sequence of partition $\{\varOmega_c: \varOmega_c=\bigcup_{k=0}^{K}\varOmega_{ck}\}_{c\geq 1}$ to approach $\varOmega$, where $\varOmega_{ck}=(\phi_k - c^{-1}, \phi_k + c^{-1}]$ for $k=1,\dots,K$ and $\varOmega_{c0}$ is the corresponding complement. Under $\varOmega_c$,
$q(G_0^{\vO_c} )= \text{d}_{K+1}\big(m_0^{-1}(G_0^{\vO_c} - M^{\vO_c})\big)$ and $p(G_0^{\vO_c} ) = \text{d}_{K+1}(G_0^{\vO_c})$, where $\text{d}_{K+1}(\cdot)$ denotes the density function for $(K+1)$-dimensional Dirichlet distribution, $M=\sum_{k=1}^{K}m_k \delta_{\phi_k}$ and $M^{\vO_c}$ is the corresponding induced random variable.
By (\ref{eq:stochastic_ELBO}), the empirical evidence lower bound under $\varOmega_c$ is
\begin{eqnarray*}
    &\ & E_{q(G_0^{\vO_c})} \Big\{\sum_{k=1}^{K} (\alpha H_k^{\vO_c} - 1)\log \frac{m_0 G_{0k}}{(G_{0k}-m_k)} 
    +  (\alpha H_0^{\vO_c} - 1)\log m_0 \\ &\ & 
   +  \sum_{s=1}^{S}  \sum_{k=1}^{K} \sum_{t=1}^{T_s} \frac{J}{S T_s} \log\frac{\Gamma(\gamma G_{0k}  +\hat{n}_{sk,t})}{\Gamma(\gamma G_{0k})}\Big\} + \text{constant},
\end{eqnarray*}
where $H_k^{\vO_c} = H(\varOmega_{ck})
$. 
Since $(G_{0k}-m_k)/m_0 \sim \text{Beta}(H_k^{\vO_c})$ under $q(G_0^{\vO_c})$, the term $ E_{q(G_0^{\vO_c})}
 (\alpha H_k^{\vO_c} - 1)\log m_0 (G_{0k}-m_k)^{-1}$ is constant with respect to parameters $\{m_k \}_{k=0}^{K}$. 
Taking $\limsup$ on both sides of the above equation with $\limsup_{\vO_c} E_{q(G_0^{\vO_c})}(\log G_{0k}) = \log m_k,$ $\limsup_{\vO_c} H_k^{\vO_c}=0$ for $k>0$ and $\limsup_{\vO_c} H_0^{\vO_c}=1,$ we obtain equation (\ref{equ:hdp_NPELBO}).

\subsection{Justification for Section \ref{sec:partition_re}}
\label{app: justification}
In this section, we show that the empirical evidence lower bound achieves the limit inferior in NPELBO. With the partition $\varOmega=(\phi_0, \phi_1, \dots, \phi_K)$ defined in Section \ref{sec:partition_re}, there is no sampling within $\phi_0$, and hence we have
\begin{equation*}
    q\big(G_0(\phi_0), G_0(\phi_1), \cdots, G_0(\phi_K)\big) \propto
    p\big(G_0(\phi_0), G_0(\phi_1), \cdots, G_0(\phi_K)\big) p\big(\bx \mid G_0(\phi_1), \cdots, G_0(\phi_K) \big).
\end{equation*}
As the likelihood part $p\big(\bx | G_0(\phi_1), \cdots, G_0(\phi_K)\big)$ does not contain $G(\phi_0)$, by integrating both sides with respect to $G_0(\phi_1), \cdots, G_0(\phi_K)$, we can get
$
    q\big(G_0(\phi_0)\big) \propto
    p\big(G_0(\phi_0)\big).
$
Moreover, the KL divergence between the variational distribution and true posterior is
\begin{eqnarray*}
    &\ &\KL\big(q(G_0^{\vO}) \parallel p(G_0^{\vO}\mid \bx)\big) \\
    &=& \int \log \frac{ q\big(G_0(\phi_0), G_0(\phi_1), \cdots, G_0(\phi_K)\big)}{p\big(G_0(\phi_0), G_0(\phi_1), \cdots, G_0(\phi_K)\big) p\big(\bx \mid G_0(\phi_1), \cdots, G_0(\phi_K)\big)} dq\big(G_0(\phi_0), G_0(\phi_1), \cdots, G_0(\phi_K)\big) \\
    &=& -\log \cN,
\end{eqnarray*}
because $$q\big(G_0(\phi_0), G_0(\phi_1), \cdots, G_0(\phi_K)\big) = p\big(G_0(\phi_0), G_0(\phi_1), \cdots, G_0(\phi_K)\big) p\big(\bx \mid G_0(\phi_1), \cdots, G_0(\phi_K)\big)/N,$$ where $N$ is the normalization constant, 
\begin{eqnarray*}
    \cN &=&\int \cdots \int p\big(G_0(\phi_0), G_0(\phi_1), \cdots, G_0(\phi_K)\big) p\big(\bx \mid G_0(\phi_1), \cdots, G_0(\phi_K)\big) dG_0(\phi_1)dG_0(\phi_1)\cdots dG_0(\phi_K)\\
    &=& \int  p\big(\bx \mid G_0(\phi_1), \cdots, G_0(\phi_K)\big) dp\big(G_0(\phi_0), G_0(\phi_1), \cdots, G_0(\phi_K)\big)\\ &=& \int  p\big(\bx \mid G_0(\phi_1), \cdots, G_0(\phi_K)\big) dp\big(G_0(\phi_1), \cdots, G_0(\phi_K)\big)
    .
\end{eqnarray*}
It is obvious that $\cN$ is independent of $p\big(G_0(\phi_0)\big)$. Therefore, if we partition $\phi_0$ into $\phi_0^{1} \cup \phi_0^{2}$, the normalization constant $\cN$ will not change, that is, the KL divergence under $\varOmega$ and $\varOmega'=(\phi_0^{1},\phi_0^{2}, \phi_1, \dots, \phi_K)$ are the same. Consequently, the partition $\varOmega$ enables the limit superior of KL divergence to be reached. By (\ref{equ:7}), the limit inferior of NPELBO is also attained.

\subsection{Derivation for (\ref{equ:hdp_gradient})}
\label{app:3}
Consider the Lagrange multiplier of constrained optimization, 
\begin{eqnarray*}
L' = - \sum_{k=1}^{K}  \log m_k 
+ (\alpha -1)\log m_0 
+ 
\sum_{s=1}^{S} \sum_{k=1}^{K} \frac{J}{S T_s} \sum_{t=1}^{T_s}\log  \frac{ \Gamma(\gamma m_k + {\hat{n}_{sk,t}})}{\Gamma(\gamma m_k) } - \lambda (\sum_{k=0}^{K} m_k -1),
\end{eqnarray*}
its first order conditions satisfy,
\begin{align*}
\begin{cases}
    JS^{-1}\gamma \sum_{s=1}^{S} \big\{ T_s^{-1} \sum_{t=1}^{T_s}\Phi(\gamma m_k +  {\hat{n}_{sk,t}})-\Phi(\gamma m_k )\big\} m_k -1 = m_k \lambda, & k=1, \dots, K,\\
    \alpha -1 = m_0 \lambda,              & k = 0.
\end{cases}
\end{align*}
Dividing $\lambda$ on both sides of the above equations, the definition of $\{m_k^* \}_{k=0}^{K}$ in (\ref{equ:hdp_gradient}) follows. 

We next show that this updating is consistent with the gradient descent after the inverse logit transformation, that is, transforming $\{m_k\}_{k=0}^{K}$ by $m_k = e^{\theta_k}/\sum_{l=0}^{K} e^{\theta_l}$ to remove the constraint of $\sum_{k=0}^{K}m_k = 1$. By $\partial m_k/\partial  \theta_k = m_k-m_k^2$, $\partial m_l/\partial \theta_k = -m_k m_l$ for $l\neq k$, and the chain rule, we have
\begin{align*}
    \frac{\partial L}{\partial \theta_k} = 
    \begin{cases}
        JS^{-1}\gamma \sum_{s=1}^{S} \big\{ T_s^{-1} \sum_{t=1}^{T_s}\Phi(\gamma m_k +  {\hat{n}_{sk,t}})-\Phi(\gamma m_k )\big\} m_k -1 - \Lambda  m_k & k=1, \dots, K,\\
        \alpha -1  - \Lambda  m_k            & k = 0,
    \end{cases}   
\end{align*}

where $L$ denotes $\widehat{\text{ELBO}}^{\vO}$ in (\ref{equ:hdp_NPELBO}) and $$\Lambda = \alpha -1 + \sum_{k=1}^{K} \Big[{JS^{-1}\gamma \sum_{s=1}^{S} \big\{ T_s^{-1} \sum_{t=1}^{T_s}\Phi(\gamma m_k +  {\hat{n}_{sk,t}})-\Phi(\gamma m_k )\big\} m_k -1}\Big].$$
As $\partial L /\partial \theta_k = \Lambda (m_k^* - m_k)$, $(m_k^* - m_k)$ represents the gradient with respect to $\theta_k$ after the inverse logit transformation.

\subsection{Derivation for (\ref{equ:general_case})}
\label{app:extensions}
For the HBNP model, we use an unnormalized random measure as the prior of $G_0.$ Given moments for Dirichlet-distributed random variables, we obtain
\begin{equation*}
\begin{split}
    \log \E_{p(G_s^\vO | G_0^\vO)} p(\hat{\bz}_{s,t}| G_s^\vO)  =
    \log \frac{\Gamma(\sum_{k=0}^{K} G_{0k} )}{\Gamma(\sum_{k=0}^{K} G_{0k} +N_s)} \prod_{k=1}^{K} \frac{\Gamma( G_{0k}  +\hat{n}_{sk,t})}{\Gamma( G_{0k})},
\end{split}
\end{equation*}
In analogy to Appendix \ref{app:2}, the empirical evidence lower bound under $\varOmega_c$ is
\begin{eqnarray*}
\begin{aligned}
K \log \mu +
 E_{q(G_0^{\vO_c})} \Big\{\sum_{k=1}^{K} \log p( G_{0k}^{\vO_c})+\log p(G_{00}^{\vO_c}) 
+  \frac{J}{S}\sum_{s=1}^{S}  \sum_{k=1}^{K}  T_s^{-1} \sum_{t=1}^{T_s} \log\frac{\Gamma(\gamma G_{0k}^{\vO_c}  +\hat{n}_{sk,t})}{\Gamma(\gamma G_{0k}^{\vO_c})}\Big\},
\end{aligned}
\end{eqnarray*}
up to a constant, 
where $K \log \mu$ comes from the Jacob matrix from $G_0, G_1,\dots, G_K$ to $\mu, m_1, \dots, m_K$. As the partition converges to single points and the corresponding complement, $\limsup _{{\vO_c}} p(G_{0k}^{\vO_c})=  v(G_{0k}^{\vO_c})$ and $\limsup _{{\vO_c}}p(G_{00}^{\vO_c}) = u(G_{00}^{\vO_c})$. Therefore, we can obtain (\ref{equ:general_case}) by  $\limsup _{{\vO_c}} G_{0k} = \mu m_k$ for $k\neq 0$ and $\limsup _{{\vO_c}} G_{00} = \mu m_0$. Specially, for the $\Gamma$DP model, $\widehat{\text{ELBO}}^{\vO}$ takes the form of
\begin{equation*}
\label{equ:gamma--Dirichlet1}
\begin{split}
\mu - \sum_{k=1}^{K} \log  m_k + (\alpha-1) \log \mu m_0 
+ \frac{J}{S}\sum_{s=1}^{S}  \bigg\{ \log \frac{\Gamma(\mu)}{\Gamma(\mu+N_s)}  +  \sum_{k=1}^{K} \frac{1}{T_s} \sum_{t=1}^{T_s}\log \frac{ \Gamma( \mu m_k + {\hat{n}_{sk,t}})}{\Gamma(\mu  m_k) }  \bigg\},
\end{split}
\end{equation*}
up to a constant 
and (\ref{eq:gamma_mu_update}) is also attained.

\section{Computational Complexity Analysis, Data and Code}
\label{ap_comp}
For CATVI, updating the global variables takes linear time, and the Monte Carlo step iteratively samples each $z_{ji}$ from $K$ possible topics. Therefore, the computational complexity of Algorithm \ref{alg:1} is dominated by $O(K + \overline{T_s}K \overline{N_S} )$, where $\overline{N_S}$ is the average number of words in a document, and $\overline{T_s}$ is the average of $T_s$ defined in Algorithm \ref{alg:1}. To implement this algorithm, we conduct our experiments on a c5d.4xlarge instance on the AWS EC2 platform, with 16 vCPUs and 32 GB RAM. It takes at most 5 hours to run all numerical experiments.

Python code for CATVI is available at
\url{https://github.com/yiruiliu110/ConditionalVI}.
We obtain the \textit{arXiv} and \textit{Wiki} data from public open resources \url{https://arxiv.org/help/bulk_data} and \url{https://dumps.wikimedia.org}, respectively. The \textit{NYT} data are from \citet{sandhaus2008}. 
For the comparison methods, we implement OVI using the Python package `gensim.models.hdpmodel' under GNU Lesser general public license v2.1. Moreover, we implement MOVI and SMVI using the Python package `bnpy' under 3-clause BSD license, which is available at \url{https://github.com/bnpy/bnpy}. Finally, the codes to run GS are available at https://github.com/linkstrife/HDP. 

\section{Sensitivity Analysis Results of CATVI}
\label{app: sensitivity}
Figure~\ref{fig:batch_size} plots the sensitivity analysis with respect to the batch size varying from 128 to 1024. Figure~\ref{fig:initial_K} plots the sensitivity analysis with respect to the initial number of topics varying from 60 to 140. We observe that the performance is not sensitive to the change of these hyperparameters.

\begin{figure}
\centering
\begin{subfigure}{0.33\linewidth}
\caption{batch size}
\label{fig:batch_size}
\includestandalone[width=1\linewidth]{hpo_batch_size}
\end{subfigure}
\begin{subfigure}{0.33\linewidth}
\caption{initial number of topics}
\label{fig:initial_K}
\includestandalone[width=1\linewidth]{hpo_initial_k}
\end{subfigure}
\caption{Plots for the perplexity vs running time for different batch sizes and initial numbers of topics.}
\label{fig:hpo}
\end{figure}

\end{document}

%% file: pictures/graphic_model.tex
\tikzset{ input/.style  = {draw, circle, node distance = 2cm}, }

\begin{tikzpicture}
\draw
node at (0,-1.5)[input, name=H]{$H$}
node [input, right of=H] (G_0) {$G_0$}
node [input, right of=G_0] (G_j) {$G_j$}
node [input, right of=G_j] (z) {$z_{ji}$}
node [input, right of=z] (x){$x_{ji}$}
node [input, above of=z] (beta){$\beta$}
node [input, left of=beta] (lambda){$\lambda$};
\draw[>=latex,->](H) -- node {}(G_0);
\draw[>=latex,->](G_0) -- node {}(G_j);
\draw[>=latex,->](G_j) -- node {}(z);
\draw[>=latex,->](z) -- node {}(x);
\draw[>=latex,->](beta) -- node {}(x);
\draw[>=latex,->](lambda) -- node {}(beta);
\draw[color=blue](3,-2.7) rectangle (9.4, -0.3);
\node at (9.4,-2.6) [above=2mm, left=0mm] {$J$};
\draw [color=red](5,-2.3) rectangle (9, -0.6);
\node at (9.1,-2.1) [above=2mm, left=0mm] {$N_j$};
\end{tikzpicture}

%% file: pictures/algo_table.tex
\begin{algorithm}
\caption{CATVI Algorithm}
\label{alg:1}
\begin{algorithmic}
\STATE{Initialize the partition $\varOmega$, the parameters for $q(G_0), q(\beta)$ and set up the step-size  $\{\rho_\tau\}_{\tau \geq 1}$.}
\REPEAT 
    \STATE{
    Randomly select $x_1, \dots, x_S$ from the dataset.}
    \FOR{$s \in \{1,\dots, S \}$} 
        \REPEAT
            \FOR{$i \in \{1,\dots, N_s\}$ }
            \STATE{
                Sample $\hat{z}_{si}\mid q(G_0)$, $q(\beta), \hat{\bz}_{s,i^-}$ 
                .
                \IF{Sampling a new $\hat{z}_{si}$}
                \STATE{Refine the  partition $\varOmega$.}
                \ENDIF}
            \ENDFOR
        \UNTIL{convergence}
        \STATE {Resample $\hat{\bz}_s=\{\hat{\bz}_{s,t}\}_{t=1}^{T_s}$.}
    \ENDFOR
    \STATE {Update parameters for $q(G_0)$ and $q(\beta)$ given samples $\{\hat{\bz}_s\}_{s=1}^S$ using the step-size $\rho_{\tau}$ 
    .}
\UNTIL {convergence}
\end{algorithmic}
\end{algorithm}

%% file: pictures/plot_1.tex
\begin{tikzpicture}
\pgfplotsset{every y tick label/.append style={font=\small, xshift=0.6ex}} 
\begin{axis}[
    xlabel={Hours},
    ylabel={Predictive perplexity},
    xmin=0, xmax=5,
    ymode=log,
    legend pos=north east,
]

\addplot [red, ultra thick] table[
    x=time,
    y=perplexity,
    col sep=comma,
    mark=none
    ]
{data/cvi_arxiv.csv};

\addplot [blue, dash pattern=on 10pt off 3pt, ultra thick] table[
    x=time,
    y=perplexity,
    col sep=comma,
    mark=none
    ]
    {data/cvi_gamma_arxiv.csv};

\addplot [loosely dotted, black, ultra thick] table[
    x=time,
    y=perplexity,
    col sep=comma,
    color=blue,
    mark=none
    ]
    {data/online_arxiv.csv};
    
\legend{CATVI--HDP,CATVI--$\Gamma$DP, OVI--HDP}
\end{axis}
\end{tikzpicture}

%% file: pictures/plot_4.tex
\begin{tikzpicture}
\pgfplotsset{every y tick label/.append style={font=\small,xshift=-0.2ex}} 
\begin{axis}[
    xlabel={Hours},
    ylabel={Number of topics},
    xmin=0, xmax=5,
    ymin=120, ymax=180,
]

\addplot [red, ultra thick] table[
    x=time,
    y=K,
    col sep=comma,
    mark=none
    ]
{data/cvi_arxiv.csv};

\addplot [blue, dash pattern=on 10pt off 3pt,ultra thick] table[
    x=time,
    y=K,
    col sep=comma,
    mark=none
    ]
    {data/cvi_gamma_arxiv.csv};

\addplot [loosely dotted, black, ultra thick] table[
    x=time,
    y=K,
    col sep=comma,
    color=blue,
    mark=none
    ]
    {data/online_arxiv.csv};
\legend{CATVI--HDP,CATVI--$\Gamma$DP, OVI--HDP}

\end{axis}
\end{tikzpicture}

%% file: pictures/plot_2.tex
\begin{tikzpicture}
\pgfplotsset{every y tick label/.append style={font=\small, xshift=0.6ex}} 

\begin{axis}[
    xlabel={Hours},
    ylabel={Predictive perplexity},
    xmin=0, xmax=5,
    ymode=log,
    legend pos=north east,
]

\addplot [red, ultra thick] table[
    x=time,
    y=perplexity,
    col sep=comma,
    mark=none
    ]
{data/cvi_nyt.csv};

\addplot [blue, dash pattern=on 10pt off 3pt, ultra thick] table[
    x=time,
    y=perplexity,
    col sep=comma,
    mark=none
    ]
    {data/cvi_gamma_nyt.csv};

\addplot [loosely dotted, black, ultra thick] table[
    x=time,
    y=perplexity,
    col sep=comma,
    color=blue,
    mark=none
    ]
    {data/online_nyt.csv};

\legend{CATVI--HDP,CATVI--$\Gamma$DP, OVI--HDP}

\end{axis}
\end{tikzpicture}

%% file: pictures/plot_5.tex
\begin{tikzpicture}
\pgfplotsset{every y tick label/.append style={font=\small,xshift=-0.2ex}} 
\begin{axis}[
    xlabel={Hours},
    ylabel={Number of topics},
    xmin=0, xmax=5,
    ymin=120, ymax=180,
]

\addplot [red, ultra thick] table[
    x=time,
    y=K,
    col sep=comma,
    mark=none
    ]
{data/cvi_nyt.csv};

\addplot [blue, dash pattern=on 10pt off 3pt, ultra thick] table[
    x=time,
    y=K,
    col sep=comma,
    mark=none
    ]
    {data/cvi_gamma_nyt.csv};

\addplot [loosely dotted, black, ultra thick] table[
    x=time,
    y=K,
    col sep=comma,
    color=blue,
    mark=none
    ]
    {data/online_nyt.csv};
\legend{CATVI--HDP,CATVI--$\Gamma$DP, OVI--HDP}

\end{axis}
\end{tikzpicture}

%% file: pictures/plot_3.tex
\begin{tikzpicture}
\pgfplotsset{every y tick label/.append style={font=\small, xshift=0.6ex}} 

\begin{axis}[
    xlabel={Hours},
    ylabel={Predictive perplexity},
    xmin=0, xmax=5,
    ymode=log,
]

\addplot [red, ultra thick] table[
    x=time,
    y=perplexity,
    col sep=comma,
    mark=none
    ]
{data/cvi_wiki.csv};

\addplot [blue, dash pattern=on 10pt off 3pt, ultra thick] table[
    x=time,
    y=perplexity,
    col sep=comma,
    mark=none
    ]
    {data/cvi_gamma_wiki.csv};

\addplot [loosely dotted, black, ultra thick] table[
    x=time,
    y=perplexity,
    col sep=comma,
    color=blue,
    mark=none
    ]
    {data/online_wiki.csv};

\legend{CATVI--HDP,CATVI--$\Gamma$DP, OVI--HDP}

\end{axis}
\end{tikzpicture}

%% file: pictures/plot_6.tex
\begin{tikzpicture}
\pgfplotsset{every y tick label/.append style={font=\small,xshift=-0.2ex}} 
\begin{axis}[
    xlabel={Hours},
    ylabel={Number of topics},
    xmin=0, xmax=5,
    legend pos=north east,
    ymin=130, ymax=240,
]

\addplot [red, ultra thick] table[
    x=time,
    y=K,
    col sep=comma,
    mark=none
    ]
{data/cvi_wiki.csv};

\addplot [blue, dash pattern=on 10pt off 3pt,ultra thick] table[
    x=time,
    y=K,
    col sep=comma,
    mark=none
    ]
    {data/cvi_gamma_wiki.csv};

\addplot [loosely dotted, black, ultra thick] table[
    x=time,
    y=K,
    col sep=comma,
    color=blue,
    mark=none
    ]
    {data/online_wiki.csv};
\legend{CATVI--HDP,CATVI--$\Gamma$DP, OVI--HDP}

\end{axis}
\end{tikzpicture}

%% file: pictures/Topics_arXiv.tex
\setlength{\tabcolsep}{0.2em}
\begin{tabular}{*{11}c}
\toprule
&                             1         & 2          & 3         & 4        & 5         & 6          & 7            & 8           & 9          & 10            \\ 
\midrule
\multirow{12}{*}{\rotatebox{90}{\textcolor{red}{CATVI}}}
&\textcolor{red}{galaxy}    & group      & network   & neutrino & \textcolor{red}{star}      & gaug       & prove        & algorithm   & collision  & test\\
& cluster   & algebra    & learn     & higg     & dwarf     & string     & bound        & optim       & product    & error\\
& redshift  & construct  & train     & matter   & survey    & brane      & theorem      & converge    & decay      & samples\\
& samples   & finit      & neural    & dark     & object    & symmetry   & finit        & solve       & hadron     & statist\\
& luminos   & lie        & image     & decay    & binari    & dimension  & class        & linear      & jet        & uncertainties\\
& formation & categories & deep      & standard & variable  & couple     & position     & approximate & transverse & fit\\
& survey    & prove      & dataset   & couple   & cluster   & action     & inequalities & gradient    & gev        & systematic\\
& agnes     & class      & feature   & mix      & stellar   & conform    & dimension    & minim       & lhc        & accuracy\\
&lar   & complex    & task      & boson    & period    & construct  & converge     & matrix      & cross      & correct\\
& \textcolor{red}{star}      & map        & object    & lepton   & photometr & correspond & continual    & iter        & section    & procedure\\
&populated & invariable & convolut       & violate       & distanc  & dual       & regular      & spars   & quark      & improve\\
& host       & manifold       & detect        & symmetry & samples    & background   & compact & constraint & collid   & bias 
\\
\midrule
\multirow{12}{*}{\rotatebox{90}{\textcolor{blue}{OVI}}} & 
\textcolor{blue}{galaxy}    & \textcolor{blue}{galaxy}     & \textcolor{blue}{star}      & \textcolor{blue}{xray}     & \textcolor{blue}{emiss}     & \textcolor{blue}{galaxy}     & higg         & \textcolor{blue}{star}        & \textcolor{blue}{emiss}      & radio          \\
& cluster   & redshift   & cluster   & \textcolor{blue}{emiss}    & \textcolor{blue}{star}      & line       & neutrino     & planet      & gammaray   & \textcolor{blue}{emiss}          \\
&halo      & source     & abundance & source   & region    & \textcolor{blue}{emiss}      & decay        & period      & source     & \textcolor{blue}{galaxy}       \\
& \textcolor{blue}{star}      & survey     & \textcolor{blue}{galaxy}    & accret   & line      & gas        & dark         & orbit       & grb        & source          \\
& stellar   & samples    & stellar   & kev      & gas       & \textcolor{blue}{star}       & boson        & dwarf       & \textcolor{blue}{xray}       & \textcolor{blue}{xray}            \\
& formation & cluster    & metal     & line     & dust      & redshift   & matter       & binari      & ray        & jet             \\
& velocity  & luminos    & age       & variable & disk      & absorption & standard     & detect      & detect     & line            \\
& dark      & agnes      & populated & spectral & molecular & samples    & couple       & stellar     & burst      & region       \\
& gas       & \textcolor{blue}{xray}       & ngc       & \textcolor{blue}{star}     & cloud     & quasar     & gev          & transit     & flux       & cluster     \\
& matter    & radio      & dwarf     & flux     & detect    & luminos    & particle     & variable    & radio      & detect        \\
& profile  & \textcolor{blue}{star}   & samples   & spectrum & formation & region    & mix      & light     & spectrum & gas       \\
& disk     & optic  & giant     & detect   & \textcolor{blue}{galaxy}    & detect    & symmetry & companion & jet      & \textcolor{blue}{star}      \\
\bottomrule
\end{tabular}

%% file: pictures/Topics_Wiki.tex
\setlength{\tabcolsep}{0.2em}
\begin{tabular}{*{11}c}
\toprule
&1         & 2          & 3         & 4        & 5         & 6          & 7            & 8           & 9          & 10  \\ 
\midrule
\multirow{12}{*}{\rotatebox{90}{\textcolor{red}{CATVI}}}    & tell   & increase     & band      & human        & james     & claim       & \textcolor{red}{armies}    & process    & polit        & album    \\
& tried  & effect       & album     & natur        & robert    & issue       & \textcolor{red}{battle}    & model      & parti        & chart     \\
& want   & case         & guitar    & tradition    & charles   & announce    & \textcolor{red}{attack}    & inform     & union        & song \\
& friend & process      & vocal     & term         & david     & critic      & \textcolor{red}{troop}     & effect     & communist    & track    \\
& leave  & measure      & track     & idea         & thomas    & controversi & \textcolor{red}{command}   & problem    & movement     & video    \\
& ask    & caus         & rock      & view         & richard   & proposal    & \textcolor{red}{soldier}   & experience & independence & billboard\\
& feel   & require      & drum      & word         & michael   & agreement   & \textcolor{red}{military}  & test       & social       & label     \\
& decide & rate         & bass      & theorie      & frank     & polit       & \textcolor{red}{fight}     & example    & republic     & peak    \\
& turn   & example      & song      & philosophies & peter     & allegation  & \textcolor{red}{tanks}     & research   & leader       & week \\
& good   & reduce       & tour      & culture      & andrew    & statement   & \textcolor{red}{brigade}   & specific   & worker       & digitated \\
& away & possibilities & studio  & believe  & brown      & agree  & \textcolor{red}{german}   & individual & socialist & hot  \\
& believe       & occur   & label    & conception & henry  & minister & \textcolor{red}{capture}    & object    & liberal    & remix   \\
\midrule
\multirow{12}{*}{\rotatebox{90}{\textcolor{blue}{OVI}}} & album  & episode      & actor     & album        & album     & episode     & \textcolor{blue}{character} & novel      & animal       & ship   \\
&band   & tell         & movi      & song         & song      & televis     & \textcolor{blue}{kill}      & character  & episode      & navies    \\
&song   & character    & character & band         & chart     & drama       & \textcolor{blue}{human}     & love       & character    & class     \\
&track  & kill         & critic    & tour         & video     & actor       & \textcolor{blue}{earth}     & poem       & voice        & boat      \\
&guitar & friend       & cast      & love         & track     & comedies    & \textcolor{blue}{attack}    & london     & movi         & naval     \\
&vocal  & leave        & review    & blue         & billboard & actress     & \textcolor{blue}{reveal}    & king       & video        & command   \\
&rock   & tried        & televis   & artist       & love      & movi        & \textcolor{blue}{episode}   & tell       & air          & vessel    \\
&tour   & relationship & episode   & rock         & label     & theatre     & \textcolor{blue}{comic}     & fiction    & dvd          & submarine \\
&chart  & need         & scene     & track        & version   & voice       & \textcolor{blue}{fight}     & narrated   & televis      & gun       \\
&studio & love         & theatre   & label        & week      & uncredit    & \textcolor{blue}{doctor}    & friend     & ray          & fleet     \\
&bass    & reveal    & picture  & chart  & peak     & nominal     & \textcolor{blue}{battle}  & mother   & blu     & sail     \\
&drum    & mother    & love     & singer & remix    & cast        & \textcolor{blue}{voice}   & critic   & song    & destroy \\ 
\bottomrule
\end{tabular}

%% file: pictures/hpo_batch_size.tex
\begin{tikzpicture}
\pgfplotsset{every y tick label/.append style={font=\small, xshift=0.6ex}} 
\begin{axis}[
    xlabel={Hours},
    ylabel={Predictive perplexity},
    xmin=0, xmax=5,
    ymax=1900,
    ymode=log,
    legend pos=north east,
]

\addplot [red, ultra thick] table[
    x=time,
    y=perplexity,
    col sep=comma,
    mark=none
    ]
{data/HPO/BATCH_SIZE/1024.csv};

\addplot [blue, dash pattern=on 10pt off 3pt, ultra thick] table[
    x=time,
    y=perplexity,
    col sep=comma,
    mark=none
    ]
    {data/HPO/BATCH_SIZE/512.csv};

\addplot [dash dot, green, ultra thick] table[
    x=time,
    y=perplexity,
    col sep=comma,
    color=blue,
    mark=none
    ]
    {data/HPO/BATCH_SIZE/256.csv};

\addplot [loosely dotted, black, ultra thick] table[
    x=time,
    y=perplexity,
    col sep=comma,
    color=blue,
    mark=none
    ]
    {data/HPO/BATCH_SIZE/128.csv};
    \legend{1024, 512, 256, 128}
  \end{axis}

\end{tikzpicture}

%% file: pictures/hpo_initial_k.tex
\begin{tikzpicture}
\pgfplotsset{every y tick label/.append style={font=\small, xshift=0.6ex}} 
\begin{axis}[
    xlabel={Hours},
    ylabel={Predictive perplexity},
    xmin=0, xmax=5,
    ymax=1900,
    ymode=log,
    legend pos=north east,
]



\addplot [red, ultra thick] table[
    x=time,
    y=perplexity,
    col sep=comma,
    color=blue,
    mark=none
    ]
    {data/HPO/INITIALK/60.csv};

\addplot [blue, dash pattern=on 10pt off 3pt, ultra thick] table[
    x=time,
    y=perplexity,
    col sep=comma,
    color=blue,
    mark=none
    ]
    {data/HPO/INITIALK/80.csv};
    
\addplot [dash dot, green, ultra thick] table[
    x=time,
    y=perplexity,
    col sep=comma,
    color=blue,
    mark=none
    ]
    {data/HPO/INITIALK/100.csv};
    
\addplot [loosely dotted, black, ultra thick] table[
    x=time,
    y=perplexity,
    col sep=comma,
    color=blue,
    mark=none
    ]
    {data/HPO/INITIALK/120.csv};
    
\addplot [brown, dash pattern=on 5pt off 2pt, ultra thick] table[
    x=time,
    y=perplexity,
    col sep=comma,
    color=blue,
    mark=none
    ]
    {data/HPO/INITIALK/140.csv};
\legend{60, 80, 100, 120, 140}
\end{axis}

\end{tikzpicture}